\documentclass[sigconf,nonacm=true,urlbreakonhyphens=true,balance=true]{acmart}
\copyrightyear{2021}
\acmYear{2021}
\setcopyright{acmcopyright}
\acmConference[ADGD '21]{Proceedings of the 1st Workshop on Synthetic Multimedia – Audiovisual Deepfake Generation and Detection}{October 24, 2021}{Virtual Event, China}
\acmBooktitle{Proceedings of the 1st Workshop on Synthetic Multimedia – Audiovisual Deepfake Generation and Detection (ADGD '21), October 24, 2021, Virtual Event, China}
\acmPrice{15.00}
\acmDOI{10.1145/3476099.3484315}
\acmISBN{978-1-4503-8682-1/21/10}

\AtBeginDocument{%
  \providecommand\BibTeX{{%
    \normalfont B\kern-0.5em{\scshape i\kern-0.25em b}\kern-0.8em\TeX}}}

\usepackage{enumitem}

\newcommand{\SystemName}{FakeAVCeleb}

\usepackage{booktabs} 
\usepackage{graphicx}
\usepackage{multirow}
\usepackage{booktabs}
\usepackage{hhline}
\usepackage{soul}
\usepackage[]{caption}
\usepackage{url}
\usepackage{xcolor}
\usepackage{color,soul}
\begin{document}

\title{Evaluation of an Audio-Video Multimodal Deepfake Dataset using Unimodal and Multimodal Detectors}

\author{Hasam Khalid, Minha Kim, Shahroz Tariq}
\affiliation{%
  \institution{College of Computing and Informatics\\ Sungkyunkwan University, South Korea}
  \city{}
  \state{}
  \country{}
}\email{{hasam.khalid, kimminha, shahroz}@g.skku.edu}

\author{Simon S. Woo}
\authornote{corresponding author}
\affiliation{%
  \institution{Department of Applied Data Science\\
Sungkyunkwan University, South Korea}
  \city{}
    \state{}
  \country{}
}\email{swoo@g.skku.edu}

\begin{abstract}
Significant advancements made in the generation of deepfakes have caused security and privacy issues. Attackers can easily impersonate a person's identity in an image by replacing his face with the target person's face. Moreover, a new domain of cloning human voices using deep-learning technologies is also emerging. Now, an attacker can generate realistic cloned voices of humans using only a few seconds of audio of the target person. With the emerging threat of potential harm deepfakes can cause, researchers have proposed deepfake detection methods. However, they only focus on detecting a single modality, i.e., either video or audio.
On the other hand, to develop a good deepfake detector that can cope with the recent advancements in deepfake generation, we need to have a detector that can detect deepfakes of multiple modalities, i.e., videos and audios. To build such a detector, we need a dataset that contains video and respective audio deepfakes. We were able to find a most recent deepfake dataset, Audio-Video Multimodal Deepfake Detection Dataset (FakeAVCeleb), that contains not only deepfake videos but synthesized fake audios as well. 
We used this multimodal deepfake dataset and performed detailed baseline experiments using state-of-the-art unimodal, ensemble-based, and multimodal detection methods to evaluate it. We conclude through detailed experimentation that unimodals, addressing only a single modality, video or audio, do not perform well compared to ensemble-based methods. Whereas purely multimodal-based baselines provide the worst performance.




\end{abstract}





\keywords{Datasets, Measurement, Deepfakes, Multimodal, Media Forensics}


\begin{teaserfigure}
\centering
  \includegraphics[width=0.8\textwidth]{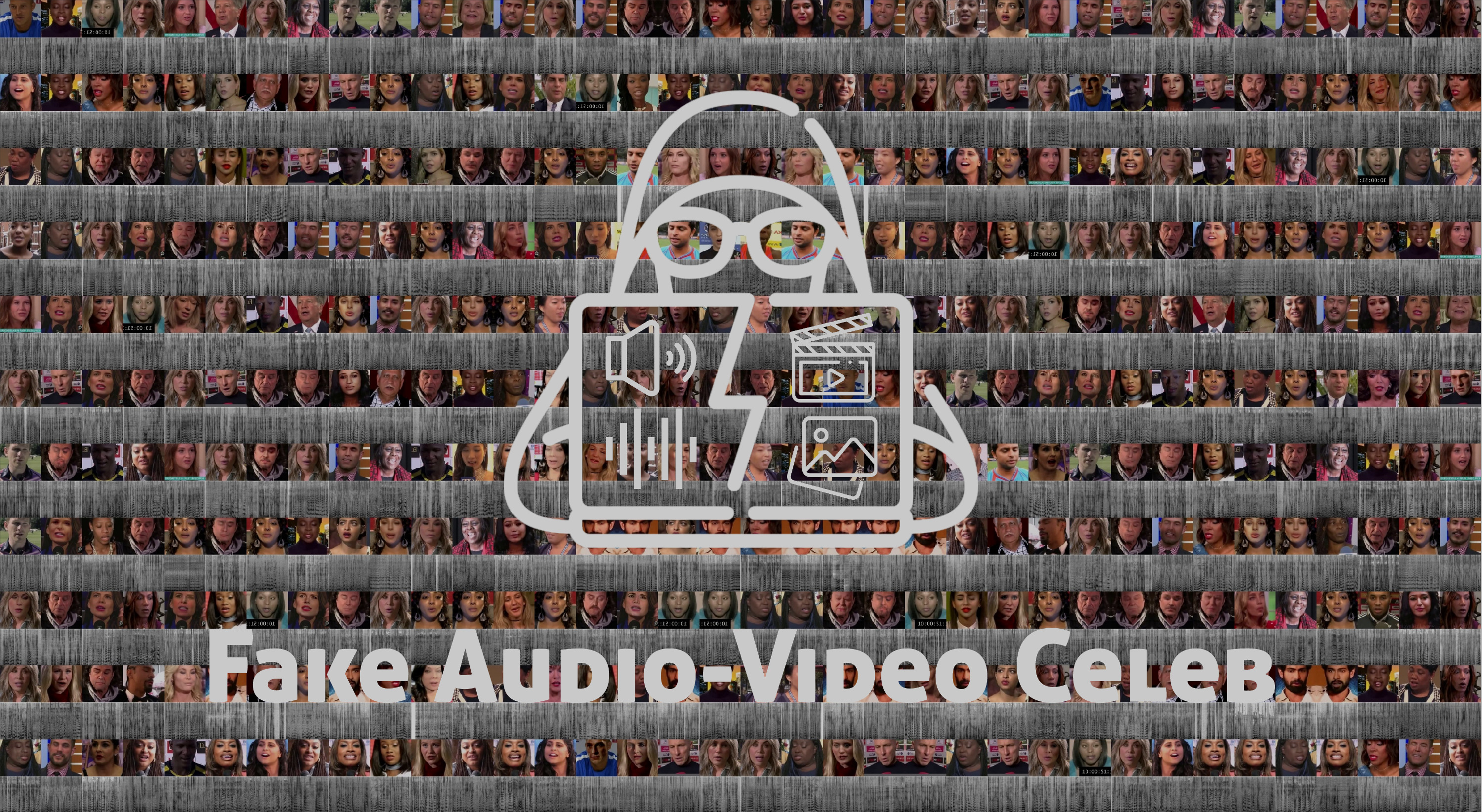}
  \caption{The Fake Audio Video Celebrity (\SystemName) dataset is one of a kind deepfake dataset. It contains three types of forgeries. 1) Real video but fake audio, 2) Real audio but fake video, and 3) Fake audio and fake video. Here, each row illustrates videos and their corresponding audio MFCCs.}
  \label{fig:teaser}
\end{teaserfigure}

\maketitle
\section{Introduction}
\label{sec:intro}
There has been a rise in forged images, videos, and audios because of the new AI technologies, particularly deep neural networks (DNNs). Image or video manipulation has been done in the past as well~\cite{sridevi2012comparative}, it has been matured by the advancements in deep neural networks since generating realistic fake human face images~\cite{goodfellow2014generative} or videos~\cite{tolosana2020deepfakes}  and cloning human voices~\cite{jia2019transfer} has become more easier and faster. Deepfakes is a technique that replaces a person's face in an image or video with another person's face using deep learning algorithms. 

Deepfakes usually combine or superimpose an existing source image onto a target image using Autoencoders (AEs)~\cite{rumelhart1985learning}, Variational Autoencoders (VAEs)~\cite{rumelhart1985learning} or Generative Adversarial Networks (GANs)\cite{goodfellow2014generative} to create a forged fake image. Sometimes AEs and GANs are exploited together to create fake images or videos.
The recent advancements in DNN-based deepfake generation methods have resulted in generating not only realistic fake images but also cloned human voices in real-time~\cite{jia2019transfer,arik2018neural}. Human voice cloning is a neural network-based speech synthesis method which takes an audio sample of the target person and a text as input and generates a high-quality speech of target speaker voice~\cite{jia2019transfer}. Example of this forgery method is a research~\cite{thies2020neural} which includes the generation of deepfake videos with very accurate lip-sync of former U.S. Presidents, e.g., Barack Obama and Donald Trump.

Therefore, the ability to easily forge videos with perfect lip-synced audios to generate deepfakes has become a well known issue~\cite{harmsofdf} and raises serious security and privacy concerns. Attackers can use deepfakes to present a forged video of a prominent person to portray false message to the public. Fake or false news has also become an issue nowadays. A huge amount of false and misleading information spread on social media platforms. False information combined with deepfake videos can be made and spread to support the agenda of an abuser to fool the public. An article was published on Forbes became a center of public attention, discussing a TV commercial ad by ESPN~\cite{forbes1}. A footage from 1998 of an ESPN analyst was shown in the video making accurate predictions about the year 2020. Later, it was found out that the clip was generated using deepfake technology~\cite{sitenew}. 

Considering the misuse and potential harm of deepfakes~\cite{ShahrozAPI}, there is vital need for deepfake detection methods to counter the misuse. A considerable amount of researchers has already contributed to this cause and proposed a numerous deepfake detection methods and techniques~\cite{nguyen2019use,yang2019exposing,zhou2017two,MesoNet,khalid2020oc,agarwal2019protecting,FakeBuster,guera2018deepfake,li2018ictu,Minha_FReTAL,Minha_CoReD}.
To build an efficient and usable deepfake detection method, a variety and huge amount of deepfake dataset is required. Hence, researchers have generated different deepfake datasets using latest deepfake generation methods~\cite{dufour2019contributing,jiang2020deeperforensics,li2020celeb,dolhansky2020deepfake,rossler2019faceforensics,kwon2021kodf}. These datasets were generated with the aim of helping researchers train and evaluate their deepfake detection methods. However, these datasets only focused on generating deepfake videos and did not consider generating fake audio respectively. 


On the other hand, Google proposed the Automatic Speaker Verification Spoofing (ASV) challenge dataset~\cite{wang2020asvspoof} with the goal of speaker verification and spoofed voice detection. However, this dataset only contains spoofed voice data and lacks the respective lip-synced or deepfake videos. Hence, the present deepfake detection datasets only covers single modality, i.e., either video or audio.This limitation of deepfake datasets also limits the deepfake detection efficiency in detecting cross domain deepfakes. As a result, existing deepfake detection methods are limited to videos or audio individually. 
As per our knowledge, there exist only one deepfake dataset, Deepfake Detection Challenge (DFDC)~\cite{dolhansky2020deepfake}, that contains a mix of deepfake video and synthesized cloned audio. But the issue with this dataset is that it is not labeled with respect to audio and video. They labeled the entire audio-video pair as fake even if one of them is real, and it is not possible to tell if the audio was fake or the video or both.

To overcome this limitation caused by the available deepfake datasets, Khalid et al.~\cite{FakeAVCeleb} proposed a novel Audio-Video Multimodal deepfake detection dataset which contains deepfake videos along with lip-synced synthesized audio. The dataset consists of videos of celebrities having different ethnic background, belonging to diverse age groups with equal proportions of men and women. We used this multimodal deepfake dataset and evaluate unimodal, ensemble-based and multimodal baseline methods. For unimodal and ensemble baselines, we used Meso-4 (and MesoInception-4~\cite{MesoNet}), Xception~\cite{XceptionNet}, EfficientNet-B0~\cite{EfficientNet}, VGG16~\cite{VGGNet} and for multimodal baselines, we used three open source multimodal methods~\cite{multi1,multi2,cdcn}. Through detailed experimentation, we conclude that the unimodal does not perform well or fails to predict deepfakes having multiple modalities. 
The main contributions of our work are summarized as follows: 
\begin{itemize}[leftmargin=10pt]

\item We performed separate (audio or video) and combined (audio + video) experiments using state-of-the-art detection models in a unimodal, ensemble, and multimodal setting for a multimodal deepfake dataset (i.e., FakeAVCeleb).

\item We demonstrate that unimodal baselines cannot detect multimodal deepfakes. Moreover, we show that the multimodal methods fail to detect deepfake from the FakeAVCeleb dataset, indicating the need for new multimodal methods for deepfakes. 


\item We find that the ensemble-based method performed the best for audio-video deepfakes. However, their prediction accuracy is not very high as well (i.e., <85\%). Therefore, concluding that multimodal deepfake detection is not a trivial task and requires further research.




\end{itemize}




\section{Related Works}
\label{sec:related}
\subsection{Fake Media Datasets}
In this section, we will briefly discuss the deepfake video and audio datasets that are publicly available.
\subsubsection{Deepfake Video Datasets}
\label{sec:dfvid}
There is a significant amount of research done in generating deepfake videos and a number of deepfake video datasets are proposed by many researchers~\cite{korshunov2018deepfakes,dolhansky2020deepfake,kwon2021kodf,li2020celeb,jiang2020deeperforensics,yang2019exposing}. The most common type of deepfake video generation is faceswap, i.e., swapping or replacing target person's face with someone else. To generate a deepfake video, Autoencoders and GANs are used most of the time and takes an excessive amount of time and computational resources~\cite{goodfellow2014generative,guera2018deepfake}. Recently, more realistic deepfake generation methods have been proposed by researchers that can generate deepfakes with better quality and takes less amount of time and computational resources~\cite{pu2021deepfake,rossler2019faceforensics,kwon2021kodf}.


The early deepfake datasets involve UADF~\cite{yang2019exposing} and Deepfake TIMIT~\cite{sanderson2009multi} and contains a small number of real videos and respective deepfake videos. 
Researchers used to propose their deepfake detection methods based on these datasets~\cite{wodajo2021deepfake,xu2021deepfake,siegel2021media}. However, the quality and quantity of these datasets are low. 
UADF contains (98) videos; meanwhile, Deepfake TIMIT contains (620) videos having real audios, respectively.

To overcome the limitations of quality and quantity, researchers proposed large-scale deepfake datasets with better quality. The most popular ones includes FaceForensics++ (FF++)~\cite{rossler2019faceforensics} and Deepfake Detection Challenge (DFDC)~\cite{dolhansky2020deepfake} dataset. The FF++ dataset contains 5,000 generated by four types of different deepfake generation methods by using a set of 1,000 real youtube videos. They later added two more types of deepfake dataset, Deepfake Detection (DFD)~\cite{rossler2019faceforensics} and FaceShifter~\cite{li2019faceshifter}. On the other hand, the DFDC dataset~\cite{dolhansky2020deepfake} was released by the collaboration of Amazon Web Services, Facebook, Microsoft, and researchers belonging to academics and released 128,154 videos captured in different environmental settings and made use of 8 types of different deepfake generation methods. The FF++ and DFDC datasets were used more often in most of the deepfake detection methods~\cite{sabir2019recurrent,agarwal2019protecting,khalid2020oc,mittal2020emotions,amerini2019deepfake}. 
However, the DFDC contains videos of people without facing camera or have extreme environmental settings, differing from real world scenarios and making it hard to detect. As per our knowledge, DFDC is the only dataset to contain respective fake audios. However, the videos are not lip-synced with the audios, and they have labeled the entire video as fake and did not specify either video or audio was fake. Furthermore, DFDC contains videos in which simply audio was replaced with some else's real audio and they have labeled it fake, which is comparatively hard to detect since both, audio and video are real.
Meanwhile, to counter the issues in deepfake datasets mentioned above, Khalid et al.~\cite{FakeAVCeleb} presented an Audio-Video Multimodal deepfake dataset containing real and fake videos of people with different ethnic backgrounds, ages, and gender. The dataset was generated using 490 videos of people with different ages, gender, and ethnicity from VoxCeleb2 dataset~\cite{chung2018voxceleb2}.

Recently, some new deepfake datasets have come into the light in which researchers have used new deepfake generation methods to generate deepfake videos. The Celeb-DF~\cite{li2020celeb} dataset was proposed in which researchers applied the modified version of the popular Faceswap method~\cite{Korshunova_2017_ICCV} on 490 YouTube real videos of 59 celebrities. Google also a proposed Deepfake Detection dataset (DFD)~\cite{rossler2019faceforensics} which contains 363 real videos and 3,000 deepfake videos, respectively. The most recent datasets includes the WildDeepfake dataset~\cite{zi2020wilddeepfake} containing 3,509 samples of some real-world deepfake videos from YouTube. KoDF~\cite{kwon2021kodf} containing 175,776 deepfake videos, and DeeperForensics-1.0~\cite{jiang2020deeperforensics} containing real videos recorded by 100 paid consensual actors and 1,000 videos from FF++ to use them as target videos and applied face swap method, resulting in 50,000 real and 10,000 fake videos. However, all of these datasets only focus on generating deepfake videos with real or no background audio, except for the FakeAVCeleb~\cite{FakeAVCeleb} which contains not only real and fake videos, but respective audios as well.

\subsubsection{Fake Audio Dataset}
Recently, generating fake or cloned human voices has become a center of attention. Researchers have proposed various methods to synthesize a human voice using DNNs. Tacotron~\cite{wang2017tacotron} is widely used for speech synthesis purposes. A few datasets have also been released by researchers. e.g., The most prominent dataset in this domain is Automatic Speaker Verification Spoofing (ASV)~\cite{wang2020asvspoof} challenge dataset, proposed by Google aiming at speaker verification and spoofed voice detection. 
DFDC also contains fake audios but it is unclear if the video is fake or the audio, as they do not provide proper labels. There exist many human voice datasets~\cite{sanderson2009multi,mccowan2005ami}, but these datasets contains real human speeches or conversations. Fake-or-Real (FoR) dataset~\cite{fakeorreal} contains a collection of more than 195,000 utterances from real humans and computer generated synthesized speech. 
However, this dataset used text-to-speech methods to just synthesize and clone a single man or a woman's voice, not of any specific person.
Moreover, in FakeAVCeleb multimodal dataset, they employed a method called Real-Time Voice Cloning (RTVC)~\cite{jia2019transfer} to generate targeted cloned human voices. Later, they used a facial reenactment method, Wav2Lip~\cite{prajwal2020lip}, to reenact the video with respect to the audio and generate a lip-synced deepfake video.

\subsection{Fake Media Detection Methods}
\subsubsection{Deepfake Video Detection}
Keeping in mind the potential misuses of deepfakes, there has been a surge in interest in deepfake detection methods. Researchers have dived into this domain and proposed DNNs based detection methods~\cite{rossler2019faceforensics,cozzolino2018forensictransfer,Shahrozpaper,khalid2020oc}. Some of the most recent methods include detection based on splice detection in which they try to exploit the inconsistencies that arise from splicing  near  the  boundaries  of  manipulated areas in an image~\cite{zhou2017two,DeepfakeDetection2,DeepfakeDetection3} abnormal eye-blinking~\cite{DeepfakeDetection7}, one-class classification~\cite{khalid2020oc,scholkopf2001estimating,oza2018one}, irregular head poses~\cite{yang2019exposing} and many other data-driven methods that do not consider particular artifact or traces~\cite{SAMTAR,Hyeonseong3,Shahroz1,Shahroz2,CLRNet,SAMGAN,khalid2020oc,transferlearning_tgd}. A transfer learning-based approach was proposed by Cozzolino et al.~\cite{cozzolino2018forensictransfer} in which they explored the possibility of generalizing a single detection method to detect deepfakes generated from multiple sources. A two-stream network was proposed by Zhou et al.~\cite{zhou2017two} for the detection of face-swapped deepfakes.

\begin{figure*}[t!]
  \includegraphics[clip, trim=15pt 15pt 15pt 15pt, width=0.8\linewidth]{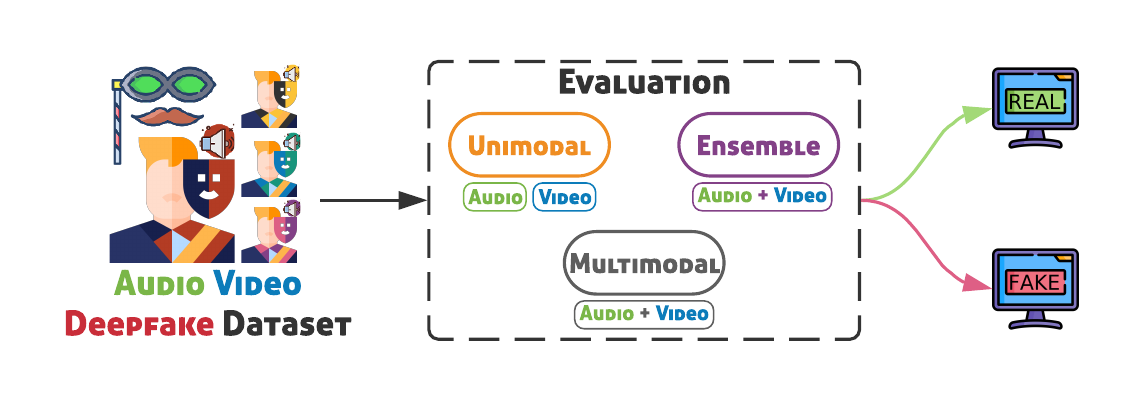}
  \caption{We used three types of detection methods to evaluate the Fake Audio-Video Celebrity Deepfake dataset. 1) Unimodal: We use a single modality, i.e., either video or audio, to train the model. 2) Ensemble: We use an ensemble of two models (one for each modality) but trained separately. 3) Multimodal: We use a single model trained on both modalities together.}
  \label{fig:pipeline}
\end{figure*}

\subsubsection{Fake Audio Detection}
Automatic speaker verification systems (ASV) can be fooled easily by spoofing attacks such as impersonation, replay (pre-recorded), or  generating synthetic audio using text-to-speech systems (TTS)~\cite{kireev2020review,wang2017tacotron,das2020predictions}.
To counter the voice spoofing attacks in speaker verification systems, researchers have proposed anti-spoofing or fake audio detection methods~\cite{patel2015combining,wang2015relative,todisco2016new}. The ASVspoof challenge series~\cite{wu2015asvspoof,todisco2019asvspoof,kinnunen2017asvspoof} have provided datasets for speaker verification and anti-spoofing detection to help researchers to develop detection methods. Traditional spoofed or fake audio detection methods gave attention to hand-crafted features such as Linear Frequency Cepstral Coefficients (LFCC)~\cite{sahidullah2015comparison}, Cochlear Filter Cepstral Coefficients Instantaneous Frequency (CFCCIF)~\cite{patel2015combining}, and Constant-Q Cepstral Coefficients
(CQCC)~\cite{todisco2016new}, combined with the Gaussian Mixture Model (GMM) as a backend-classifier~\cite{wang2015relative,sanchez2015toward,sahidullah2015comparison,patel2015combining}. Deep learning-based fake audio detection methods have also been proposed by researchers in which they used Convolutional Neural Networks (CNNs),  Recurrent Neural Networks (RNNs), and the combination of CNN and RNN~\cite{zhang2017investigation,monteiro2020generalized,monteiro2020generalized}. To train the deep learning models, feature extraction methods such as fast Fourier transform (FFT)~\cite{nussbaumer1981fast} and Mel-frequency cepstrum (MFCC)~\cite{hasan2004speaker} are applied to convert raw audio files into feature vectors before passing them to the model.

\subsubsection{Multi-modal Detection}
Multimodal learning refers to an embodied learning situation which includes learning multiple modalities such as audio, video, text, etc. Recently, there has been some recent interest in multimodal learning because of the fact that multimodal outperforms single feature-based methods~\cite{mcgiff2019towards,kachele2014fusion,meng2013depression,ortega2019multimodal}. 
Multimodal learning methods have been applied to various problems involving human activity such as action recognition~\cite{memmesheimer2020gimme}, gesture recognition~\cite{neverova2015moddrop}, face recognition~\cite{ding2015robust}, gaze-direction estimation~\cite{mukherjee2015deep}, and emotion recognition.
Much of the work in multimodal learning includes recognition
and prediction problems associated with representations of text, audio, video, or image data~\cite{epstein2018joint,vukotic2016bidirectional,chandar2016correlational}. 
Mittal et al. proposed an emotions based audio-visual deepfake detection method~\cite{mittal2020emotions} inspired by Siamese network, in which they detect a deepfake video based on perceived emotion from the two modalities (audio and video) within a video. Chugh et al. also proposed deepfake video detector~\cite{chugh2020not} that detects a deepfake video based on the dissimilarity between the audio and visual modalities
However, both of them used DFDC and DeepFake-TIMIT which makes the multimodal deepfake detection not efficient because of the deficiencies of the datasets we mentioned in section~\ref{sec:dfvid}.
In this paper, we are particularly focused on deepfake video and fake audio detection problems. As per our knowledge, 
the deepfake video and audio detection simultaneously using multimodal learning has not been explored before.

\section{Multimodal Deepfake Detection}
\label{sec:method}


In this section, we provide the details of multimodal deepfake dataset, FakeAVCeleb, and the detection methods that we applied on this dataset.

\subsection{Dataset}
To evaluate our baselines and to perform analysis of the effects of multimodal deepfake dataset on detection methods, we used an audio-video multimodal deepfake detection dataset (i.e., FakeAVCeleb~\cite{FakeAVCeleb}).
The dataset was generated using real videos from the VoxCeleb2~\cite{chung2018voxceleb2} dataset. 
It is gender and racial unbiased since the real video set belongs to the celebrities having five different ethnic backgrounds, Caucasian (Americans), Caucasian (Europeans), Black (African), South Asian (Indian), and East Asian (e.g., Chinese, Korean, and Japanese). Each ethnic group consists of 100 real videos of 100 celebrities, 50 for each gender, resulting in 600 unique videos with an average of 7 seconds duration. Moreover, the dataset is not public yet so we requested the authors to get the dataset.


To generate the dataset, they used different deepfake generation methods, including face-swapping and facial reenactment methods. For the fake audio generation, they used synthetic speech synthesis methods to generate cloned or fake voice samples of the people in the videos. To generate deepfake videos, we used Faceswap~\cite{Korshunova_2017_ICCV}, Faceswap GAN (FSGAN)~\cite{nirkin2019fsgan}, and DeepFaceLab~\cite{Korshunova_2017_ICCV} to perform face swaps. On the other hand, they used the Real-Time Voice Cloning tool (RTVC)~\cite{jia2019transfer} to generate fake audios.
Later, they applied Wav2Lip for facial reenactment based on source audio.



\subsection{Evaluation Methods}
We perform frame-based deepfake detection and classify each extracted video frame as real or fake.
We evaluate the FakeAVCeleb multimodal dataset with three different evaluation methods. 1) Unimodal, 2) Ensemble and 3) Multimodal. We will cover their details in the following sections.
\subsubsection{Unimodal Methods}
In a unimodal setting, we evaluate only one modality at a time (i.e., audio or video; see Fig.~\ref{fig:pipeline} for a visual illustration). We explore several baselines to observe their their effects on the multimodal dataset. The following is a brief description of them.
\begin{enumerate}[leftmargin=14pt]
    \item \textit{\textbf{VGG16}}~\cite{VGGNet} consists of small convolution filters of size $3\times3$, which show improvement of detection accuracy, significantly. Moreover, VGG is considered a standard baseline for many image classification tasks. Therefore, it becomes a natural baseline option.
    \item \textit{\textbf{Meso-4}} is proposed by Afchar et al.~\cite{MesoNet} for detecting face tempering in videos. Meso-4 has a small number of layers and focuses on mesoscopic (middle-level) features.
    \item \textit{\textbf{MesoInception-4}} is also proposed by Afchar et al.~\cite{MesoNet}. The model architecture is inspired by InceptionNet~\cite{InceptionNet} to detect fake videos and facial manipulations effectively. It improves Meso-4 by replacing the first convolutional layers with modules of InceptionNet.
    \item \textit{\textbf{Xception}}~\cite{XceptionNet} is considered the state-of-the-art deep learning model for deepfake detection tasks. Xception applies depth-wise separable convolutions, and Rossler et al.~\cite{rossler2019faceforensics} demonstrated that the Xception achieves the best accuracy on the FaceForensics++ dataset.
    \item \textit{\textbf{EfficientNet-B0}}~\cite{EfficientNet} is proposed with efficient composite coefficients to scale all depth, width, and resolution dimensions uniformly. This method shows high accuracy with an order of magnitude fewer parameters.
\end{enumerate}
One major drawback of the unimodal method is that it cannot consider both modalities together. Therefore it will fail in scenarios where each modality has different label. For example, if audio is fake, but the video is real, the unimodal classifier trained on audio (video) will predict it as fake (real) and vice versa. Therefore, methods that can consider both modalities are significantly important for audio-video deepfake detection.

\subsubsection{Ensemble Methods}
The ensemble-based methods are used to boost the prediction performance by combining two or more classifiers. In our ensemble method settings, we use one model (classifier) trained on audio (MFCCs) and one on video (frames). The voting method is an easy way to incorporate prediction from several classifiers. Therefore, to evaluate the performance of video with audio simultaneously, we use two voting methods: 1) soft-voting and 2) hard-voting ensembles. Soft-voting uses a class-wise average of probabilities from each classifier, whereas hard-voting employs a majority vote from the probabilities of models. We use the baseline method from the unimodal method to build the ensembles.

\subsubsection{Multimodal Methods}
\label{multidetail1}
For multimodal evaluation of the baselines on FakeAVCeleb multimodal dataset, we used three different publicly available multimodal methods. 

\begin{enumerate}[leftmargin=14pt]
    \item \textbf{\textit{Multimodal-1}}~\cite{multi1} The first multimodal classification method consists of a combination of 3 neural networks, 1) extract features from visual modality 2) extracts features from textual modality, 3) decides which one of the two modalities is more informative, and then performs classification by paying attention to the more informative modality. They used their model to perform the classification of food recipes.

    \item \textbf{\textit{Multimodal-2}}~\cite{multi2} For the second model, we used an open-source multimodal implementation for movie genre prediction, which takes movie posters and genre as input. The model architecture consists of three blocks, a CNN block for a movie poster (visual modality), an LSTM block for movie genre (textual modality), and a feed-forward network for classification that takes the concatenated output from the first two blocks.
    
    \item \textbf{\textit{CDCN}}~\cite{cdcn} The third multimodal is based on Central Difference Convolutional Networks (CDCN)~\cite{cdcn} used to solve the task of face anti-spoofing. The model takes three-level fused features (low-level, mid-level, high-level) extracted for predicting grayscale facial depth. 
    
\end{enumerate}

\section{Experiments and Results}
\label{sec:exp}


\begin{table*}[t!]
\centering
\caption{Performance of real and fake video and audio on five \textit{unimodal} baselines, respectively. The highest performance values are shown in bold.}
\label{table:performance_single}
\resizebox{\linewidth}{!}{%
\begin{tabular}{l|c|cccc|cccc} 
\toprule
\multirow{2}{*}{\textbf{Model}} & \multirow{2}{*}{\textbf{Class}} & \multicolumn{4}{c|}{\textbf{Unimodal (Video)}} & \multicolumn{4}{c}{\textbf{Unimodal (Audio)}} \\ 
\cline{3-10}
 &  & \textbf{\ Precision\ } & \textbf{\ Recall\ } & \textbf{\ F$_1$-score\ } & \textbf{\ Accuracy\ } & \textbf{\ Precision\ } & \textbf{\ Recall\ } & \textbf{\ F$_1$-score\ } & \textbf{\ Accuracy\ } \\ 
\hline\hline
\multirow{2}{*}{\textbf{MesoInception-4}~\cite{MesoNet}} & Real & \textbf{0.7538} & 0.7329 & 0.7432 & \multirow{2}{*}{0.7788} & \textbf{1.0000} & 0.0725 & 0.1351 & \multirow{2}{*}{0.5396} \\
 & Fake & 0.7972 & \textbf{0.8143} & 0.8056 &  & 0.5224 & \textbf{1.0000} & 0.6863 &  \\ 
\hline
\multirow{2}{*}{\textbf{Meso-4}~\cite{MesoNet}} & Real & 0.4329 & \textbf{0.9723} & 0.5991 & \multirow{2}{*}{0.4315} & 0.5000 & 0.8841 & 0.6387 & \multirow{2}{*}{0.5036} \\
 & Fake & 0.3580 & 0.0120 & 0.0232 &  & 0.5294 & 0.1286 & 0.2069 &  \\ 
\hline
\multirow{2}{*}{\textbf{Xception}~\cite{XceptionNet}} & Real & 0.6654 & 0.7708 & 0.7143 & \multirow{2}{*}{0.7306} & 0.8750 & 0.6087 & 0.7179 & \multirow{2}{*}{\textbf{0.7626}} \\
 & Fake & \textbf{0.7973} & 0.6993 & 0.7451 &  & \textbf{0.7033} & 0.9143 & \textbf{0.7950} &  \\ 
\hline
\multirow{2}{*}{\textbf{EfficientNet-B0}~\cite{EfficientNet}} & Real & 0.5254 & 0.7873 & 0.6303 & \multirow{2}{*}{0.5964} & 0.5000 & \textbf{1.0000} & 0.6667 & \multirow{2}{*}{0.5000} \\
 & Fake & 0.7310 & 0.4483 & 0.5558 &  & 0.0000 & 0.0000 & 0.0000 &  \\ 
\hline
\multirow{2}{*}{\textbf{VGG16}~\cite{VGGNet}} & Real & 0.6935 & 0.8966 & \textbf{0.7821} & \multirow{2}{*}{\textbf{0.8103}} & 0.8948 & 0.6894 & \textbf{0.7788} & \multirow{2}{*}{0.6714} \\
 & Fake & 0.8724 & 0.7750 & \textbf{0.8208} &  & 0.6200 & 0.8857 & 0.7294 &  \\
\bottomrule
\end{tabular}
}
\end{table*}

\subsection{Preprocessing}
To perform experimentation, we performed some preprocessing steps on FakeAVCeleb multimodal dataset. Both modalities, videos, and audios were preprocessed separately.
Since the source of the real videos of FakeAVCeleb multimodal dataset is VoxCeleb2 dataset, the videos are already face-centered and cropped. We extract respective frames from each video and store them separately. For the audio modality, we extract audio from the videos and store them in a WAV format with a sampling rate of 16 kHz. These audio files are in raw format, so before giving audio directly to the model, we first compute Mel-Frequency Cepstral Coefficients (MFCC) features by applying a $25ms$ Hann window~\cite{enwiki} with $10ms$ window shifts, followed by a fast Fourier transform (FFT) with $512$ points. As a result, we obtain a 2D array of $80$ MFCC features $(D = 80)$ per audio frame and store the resulting MFCC features as a three-channel image. These MFCC images are then passed to the model as an input to extract speech features to learn the difference between real and fake human speeches. To evaluate all three types of baseline methods, unimodal, ensemble, and multimodal, we used FakeAVCeleb to build a general test set that contains real and deepfake videos. The test set contains 70 real and 70 fake videos. The videos belongs to the individuals not in training set so that it would not have any bias in the results.
By applying the preprocessing steps mentioned above, we were able to obtain 1,895 real and 2,418 fake frames for videos, and 70 real and 70 fake MFCC feature images for audios, respectively.

\subsection{Experiment Settings}
We train each method for 50 iterations using early stopping with a patience value of 10. We employ Adam optimizer with a learning rate of $1\times 10^{-5}$. Our experiments are run on an Intel Core i7-9700 CPU with 32 GB of RAM and Nvidia RTX 3090 GPU. We use Precision, recall, F$_1$-score, and accuracy metric for evaluation.

The following sections will provide the experimental settings for unimodal, ensemble-based, and multimodal methods.

\begin{table*}
\centering
\caption{Performance of real and fake video with audio using Soft-voting method and Hard-voting method on five baselines. The highest performance values are shown in bold.}
\label{table:performance_ensemble}
\resizebox{\linewidth}{!}{%
\begin{tabular}{l|c|cccc|cccc} 
\toprule
\multirow{2}{*}{\textbf{Model}} & \multirow{2}{*}{\textbf{Class}} & \multicolumn{4}{c|}{\textbf{Ensemble Soft-voting (Audio $+$ Video)}} & \multicolumn{4}{c}{\textbf{Ensemble Hard-voting (Audio $+$ Video)}} \\ 
\cline{3-10}
 &  & \multicolumn{1}{l}{\textbf{\ Precision\ }} & \multicolumn{1}{l}{\textbf{\ Recall\ }} & \multicolumn{1}{l}{\textbf{\ F$_1$-score\ }} & \textbf{\ Accuracy\ } & \multicolumn{1}{l}{\textbf{\ Precision\ }} & \multicolumn{1}{l}{\textbf{\ Recall\ }} & \multicolumn{1}{l}{\textbf{\ F$_1$-score\ }} & \textbf{\ Accuracy\ } \\ 
\hline\hline
\multirow{2}{*}{\textbf{MesoInception-4}~\cite{MesoNet}} & Real & 0.6451 & 0.8507 & 0.7337 & \multirow{2}{*}{0.7287} & 0.6451 & 0.8507 & 0.7337 & \multirow{2}{*}{0.7287} \\
 & Fake & 0.8440 & 0.6332 & 0.7235 &  & 0.8440 & 0.6332 & 0.7235 &  \\ 
\hline
\multirow{2}{*}{\textbf{Meso-4}~\cite{MesoNet}} & Real & 0.4451 & 0.9351 & 0.6031 & \multirow{2}{*}{0.4593} & 0.4451 & 0.9351 & 0.6031 & \multirow{2}{*}{0.4593} \\
 & Fake & 0.6295 & 0.0864 & 0.1520 &  & 0.6295 & 0.0864 & 0.1520 &  \\ 
\hline
\multirow{2}{*}{\textbf{XceptionNet}~\cite{XceptionNet}} & Real & 0.4394 & \textbf{1.0000} & 0.6105 & \multirow{2}{*}{0.4394} & 0.4394 & \textbf{1.0000} & 0.6105 & \multirow{2}{*}{0.4394} \\
 & Fake & 0.0000 & 0.0000 & 0.0000 &  & 0.0000 & 0.0000 & 0.0000 &  \\ 
\hline
\multirow{2}{*}{\textbf{EfficientNet-B0}~\cite{EfficientNet}} & Real & 0.5630 & 0.7235 & 0.6333 & \multirow{2}{*}{0.6318} & 0.5630 & 0.7235 & 0.6333 & \multirow{2}{*}{0.6318} \\
 & Fake & 0.7210 & 0.5600 & 0.6304 &  & 0.7210 & 0.5600 & 0.6304 &  \\ 
\hline
\multirow{2}{*}{\textbf{VGG16}~\cite{VGGNet}} & Real & \textbf{0.6935} & 0.8966 & \textbf{0.7821} & \multirow{2}{*}{\textbf{0.7804}} & \textbf{0.6935} & 0.8966 & \textbf{0.7821} & \multirow{2}{*}{\textbf{0.7804}} \\
 & Fake & \textbf{0.8948} & \textbf{0.6894} & \textbf{0.7788} &  & \textbf{0.8948} & \textbf{0.6894} & \textbf{0.7788} &  \\
\bottomrule
\end{tabular}
}
\end{table*}

\subsubsection{Unimodal}

We used FakeAVCeleb to perform experiments using five baseline models, that are, VGG16~\cite{VGGNet}, MesoInceptation-4, Meso-4~\cite{MesoNet}, Xception~\cite{XceptionNet}, and EfficientNet-B0~\cite{EfficientNet} in a unimodal setting. We train these unimodal classifiers using 133,724 frames extracted from 360 fake and 360 real videos for the video-only setting. And, we train the classifiers using 720 MFCCs generated from the audio of the same 360 fake and 360 real videos for the audio-only setting.


    

\subsubsection{Ensemble}

To evaluate the multiple modalities of FakeAVCeleb multimodal dataset, we use an ensemble of one audio and one video classifier using hard- and soft-voting. We use the same baselines from the unimodal 
but now in a two classifiers setting, one for each modality. We sample 25 frames from each video; however, we have only one MFCC features image per video. Therefore, we upsample the number of MFCC feature images to balance them with the number of frames to be trained and tested in pairs.  Note: \textit{It is possible to handle the imbalance using other methods. For example, we can generate multiple MFCC feature images from the audio by dividing it into shorter durations. However, it is out of this paper's scope.}


\subsubsection{Multimodal Audio+Video}
As mentioned in section~\ref{multidetail1}, to observe the effects of FakeAVCeleb multimodal dataset, we used three different multimodal classification methods, Multimodal classification for food recipes (Multimodal-1), Multimodal classification for movie genre prediction (Multimodal-2), and Central Difference Convolutional Networks (CDCN) for face anti-spoofing. 
Even though these methods are not originally built for the task of deepfake detection, we chose these methods as we could not find any multimodal method that can detect an audio-visual deepfake.
Since these models were built to perform certain classification and prediction tasks and our preprocessed dataset consists of 2 visual modalities, video frames and audio MFCC features, we modified these models with respect to the video and audio modalities of FakeAVCeleb multimodal dataset. For the Multimodal-1 method, we removed the neural network for textual modality and replicated the same neural network for a visual modality to use this model on the multimodal dataset. For the Multimodal-2 method, we removed the LSTM block and replaced it with the similar CNN block, resulting in two CNN blocks, one for visual and one for audio modality. Moreover, for the third multimodal, CDCN, we modified the model according to our need by removing the third modality since it contains all three visual modalities.

\subsection{Unimodal Results}
In this section, we report how the baseline trained on audio or video performed on the test set, which contains real and all three types of fakes from the FakeAVCeleb dataset. Table~\ref{table:performance_single} presents the results of deepfake detection for audio and video separately using unimodal baselines. We can observe that the best detection performance is around 76\% for audio and 81\% for video.

\subsubsection{Results for Video-only Trained Classifier}
For video, VGG (81\%) and Meso-4 (43\%) demonstrate the best and worst performance accuracy, respectively. In particular, the recall score of Meso-4 indicates that the model fails to detect most fake videos. It is unexpected and interesting that VGG16 outperformed Xception on this task, given that Xception is the best performer on other deepfake datasets such as FaceForensics++. However, we assume that with proper hyper-parameter tuning, the Xception model might outperform VGG, but we have not considered hyper-parameter tuning for any of the baseline methods. Therefore, it is out of our scope.

\subsubsection{Results for Audio-only Trained Classifier}
For audio, Xception (76\%) shows the best prediction performance, and EfficientNet-B0 (50\%) shows the worst prediction performance. It is interesting to note that Meso-4 overfits real class and MesoInception-4 overfits fake class for audio detection. We can observe that none of the baselines provides satisfactory detection performance, indicating that SOTA deepfake detection methods are not suitable for fake audio detection. The models developed for human speech verification or detection may perform better for fake audio detection. However, we have not considered such methods in this work. Nevertheless, future work can explore in this direction.

\subsubsection{Summary}
Overall, Xception is the most stable performer (Audio: 76\% and Video: 73\%). Generally, the results from Table~\ref{table:performance_single} indicate that it is relatively difficult for the state-of-the-art deepfake detection model to distinguish between real and fake audio and video from FakeAVCeleb dataset.

\subsection{Ensemble Results}
In Table~\ref{table:performance_ensemble}, we present the results from an ensemble of one unimodal audio and one unimodal video classifier. The ensemble of VGG classifiers performs the best (78\%), and Xception classifiers perform the worst (43\%) on the test set. MesoInception-4 ensemble demonstrates the 2nd best performance (72\%).

Overall, we can observe that choice of soft- or hard-voting had no significant impact on the performance of the ensemble classifier, as they show the same prediction score (see Table~\ref{table:performance_ensemble}). Note: \textit{This is because we have only two classifiers in our ensemble}. Moreover, none of the ensemble-based methods could achieve a high detection score (i.e., >90\%), demonstrating that detecting multimodal audio-video deepfakes is not a trivial task, and better detection methods are required.

\subsection{Multimodal Results}
In this section, we report how the three baseline multimodals performed on the two modalities, video and audio, of the FakeAVCeleb multimodal dataset. We trained the Multimodal-1 for 50 epochs. After selecting the best performing epoch, the model was able to classify with $50\%$ accuracy. For the Multimodal-2, we trained it for 50 epochs and evaluated it on the dataset, giving us $67.3\%$ accuracy.
The third model, CDCN, was also trained for 50 epochs, giving us around $51\%$ accuracy. The precision, recall, F$_1$-score, and accuracy of each model are reported in Table~\ref{table:performance_multimodal}. We can observe that the Multimodal-1 and CDCN did not perform well as compared to the Multimodal-2. One possible reason for this behavior is that these models are built to perform a specific task, i.e., food recipe classification and movie genre prediction. Furthermore, it is challenging for multimodal methods to detect a deepfake in which either video is fake or the audio. Therefore, more research is needed in developing better multimodal deepfake detectors.


\begin{table}[t!]
\centering
\caption{Performance of real and fake video with audio three different open source \textit{multimodal} methods, respectively. The highest performance values are shown in bold.}
\label{table:performance_multimodal}
\resizebox{\linewidth}{!}{%
\begin{tabular}{l|c|cccc} 
\toprule
\multirow{2}{*}{\textbf{Model}}  & \multirow{2}{*}{\textbf{Class}} & \multicolumn{4}{c}{\textbf{Multimodal (Audio $+$ Video)}}  \\ 
\cline{3-6}
&  & \multicolumn{1}{c}{\textbf{Pre.}} & \multicolumn{1}{c}{\textbf{Rec.}} & \multicolumn{1}{c}{\textbf{F$_1$-score}} & \textbf{Acc.}                     \\ 
\hline\hline
\multirow{2}{*}{\textbf{Multimodal-1}~\cite{multi1}}  & Real                   & 0.000 & 0.000 & 0.000 & \multirow{2}{*}{0.500} \\
& Fake & 0.496  &  \textbf{1.000}  & 0.663   \\ 
\hline
\multirow{2}{*}{\textbf{Multimodal-2}~\cite{multi2}} & Real & \textbf{0.710 } &\textbf{0.587}  & \textbf{0.643}  & \multirow{2}{*}{\textbf{0.674}}  \\
& Fake & \textbf{0.648}  & 0.760  & \textbf{0.700}  \\ 
\hline
\multirow{2}{*}{\textbf{CDCN}~\cite{cdcn}}  & Real & 0.500 & 0.068 & 0.120 & \multirow{2}{*}{0.515} \\
& Fake  & 0.500   & 0.940 & 0.651    \\

\bottomrule
\end{tabular}
}
\end{table}

\section{Conclusion}
\label{sec:conclusion}

With the advent of new deepfake generation methods that can not only generate realistic deepfake videos but perfect lip-synced cloned human voice as well, the detection of the deepfake having two different modalities is a new challenging task. In this paper, we performed various experimentation and discuss the effects of multimodal deepfakes on existing state-of-the-art detection methods. We used a recent multimodal deepfake dataset (FakeAVCeleb) and evaluate the detection methods belonging to three categories, unimodal, ensemble-based, and multimodal.
After a detailed analysis of the detection performance, we conclude that the performance of unimodals, focusing only on a single modality, video or audio, does not perform well compared to ensemble-based methods. Multimodal methods, on the other hand, performs the worst.

\begin{acks}
This work was partly supported by Institute of Information \& communications Technology Planning \& Evaluation (IITP) grant funded by the Korea government (MSIT) (No.2019-0-00421, AI Graduate School Support Program (Sungkyunkwan University)), (No. 2019-0-01343, Regional strategic industry convergence security core talent training business) and the Basic Science Research Program through National Research Foundation of Korea (NRF) grant funded by Korea government MSIT (No. 2020R1C1C1006004). Also, this research was partly supported by IITP grant funded by the Korea government MSIT (No. 2021-0-00017, Original Technology Development of Artificial Intelligence Industry), and was partly supported by the MSIT (Ministry of Science, ICT), Korea, under the High-Potential Individuals Global Training Program (2020-0-01550) supervised by the IITP (Institute for Information \& Communications Technology Planning \& Evaluation).
\end{acks}




\bibliographystyle{ACM-Reference-Format}
\balance
\bibliography{acmart.bib}

\end{document}